\documentclass[runningheads]{llncs}

\usepackage{graphicx}
\usepackage{comment}
\usepackage{amsmath,amssymb}
\usepackage{color}
\usepackage{url}
\usepackage{hyperref}
\usepackage{bm}
\usepackage{xcolor}
\usepackage[export]{adjustbox}
\usepackage{multirow}
\usepackage{booktabs}
\usepackage[utf8]{inputenc}
\usepackage{caption}
\usepackage{subcaption}

\newif\ifreview

\ifreview
	\usepackage{lineno}

	\linenumbers
\fi

\begin{document}

\newcommand{\etal}{\textit{et al.\.}}


\def\SubNumber{067}

\def\GCPRTrack{Pattern Recognition in the Life and Natural Sciences Track}

\title{Improving Data Efficiency for Plant Cover Prediction with Label Interpolation and Monte-Carlo Cropping}
\titlerunning{Improving Data Efficiency for Plant Cover Prediction}

\ifreview
\else

	\author{
		Matthias Körschens\inst{1,2}\orcidID{0000-0002-0755-2006} \and
		Solveig Franziska Bucher\inst{1,2,3}\orcidID{0000-0002-2303-4583} \and
		Christine Römermann\inst{1,2,3}\orcidID{0000-0003-3471-0951} \and
		Joachim Denzler\inst{1,2,3}\orcidID{0000-0002-3193-3300}
	}

	\authorrunning{M. Körschens et al.}

	\institute{
		Friedrich Schiller University, D-07743 Jena, Germany \\ \and
		German Centre for Integrative Biodiversity Research (iDiv) Halle-Jena-Leipzig, D-04103 Leipzig, Germany \\ \and
		Michael Stifel Center Jena, D-07743 Jena, Germany \\
	\email{\{matthias.koerschens,solveig.franziska.bucher,\\christine.roemermann,joachim.denzler\}@uni-jena.de}
	}
\fi

\maketitle              

\begin{abstract}
	The plant community composition is an essential indicator of environmental changes and is, for this reason, usually analyzed in ecological field studies in terms of the so-called plant cover. The manual acquisition of this kind of data is time-consuming, laborious, and prone to human error. Automated camera systems can collect high-resolution images of the surveyed vegetation plots at a high frequency. In combination with subsequent algorithmic analysis, it is possible to objectively extract information on plant community composition quickly and with little human effort.
	An automated camera system can easily collect the large amounts of image data necessary to train a Deep Learning system for automatic analysis. However, due to the amount of work required to annotate vegetation images with plant cover data, only few labeled samples are available. As automated camera systems can collect many pictures without labels, we introduce an approach to interpolate the sparse labels in the collected vegetation plot time series down to the intermediate dense and unlabeled images to artificially increase our training dataset to seven times its original size.
	Moreover, we introduce a new method we call Monte-Carlo Cropping. This approach trains on a collection of cropped parts of the training images to deal with high-resolution images efficiently, implicitly augment the training images, and speed up training.
	We evaluate both approaches on a plant cover dataset containing images of herbaceous plant communities and find that our methods lead to improvements in the species, community, and segmentation metrics investigated. 



\end{abstract}
\section{Introduction}

\newcommand{\ann}[3]{
	\centering
	\footnotesize
	\begin{tabular}{lr}
		\textit{Trifolium pra.} & #1 \\
		\textit{Achillea mil.} & #2 \\
		\textit{Grasses} & #3 \\
		... & ... \\
	\end{tabular} \\
}
\newcommand{\noann}{\ann{?}{?}{?}}

\begin{figure}[t]
	\centering
	\adjustbox{cfbox=green 1pt 1pt 0pt, minipage=0.23\textwidth}{
		\centering
		\includegraphics[width=\textwidth]{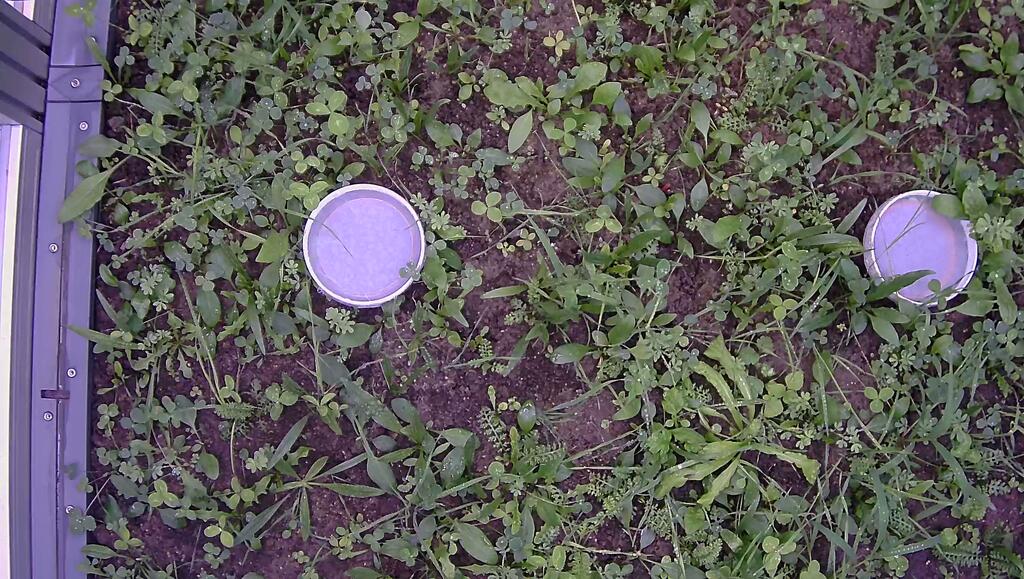}\\
		(0)\\
		Annotated\\[0.5em]
		\ann{15\%}{5\%}{25\%}
	}
	\adjustbox{cfbox=red 1pt 1pt 0pt, minipage=0.23\textwidth}{
		\centering
		\includegraphics[width=\textwidth]{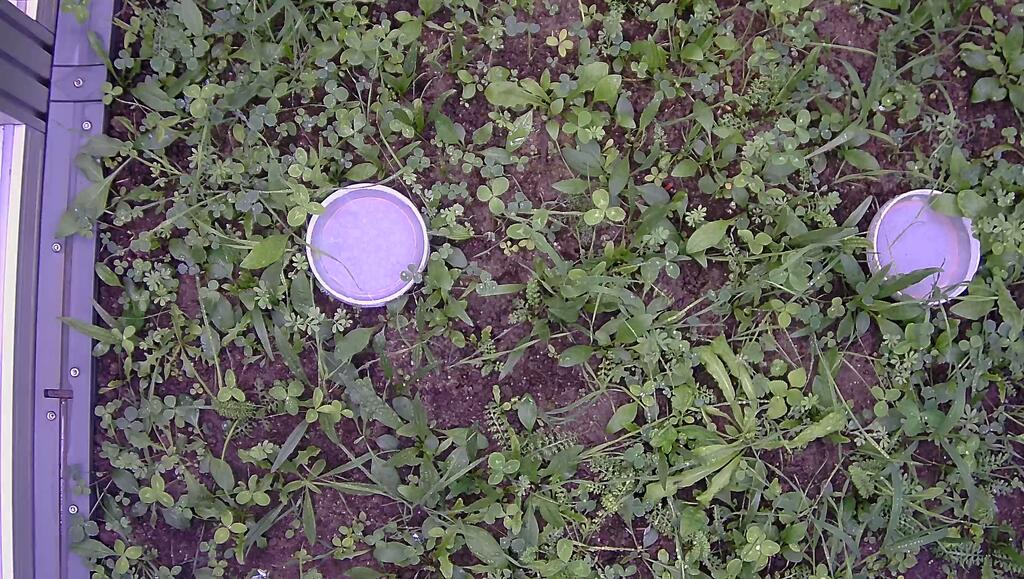}\\
		(1)\\
		Not Annotated\\[0.5em]
		\noann{}
	}
	\adjustbox{cfbox=red 1pt 1pt 0pt, minipage=0.23\textwidth}{
		\centering
		\includegraphics[width=\textwidth]{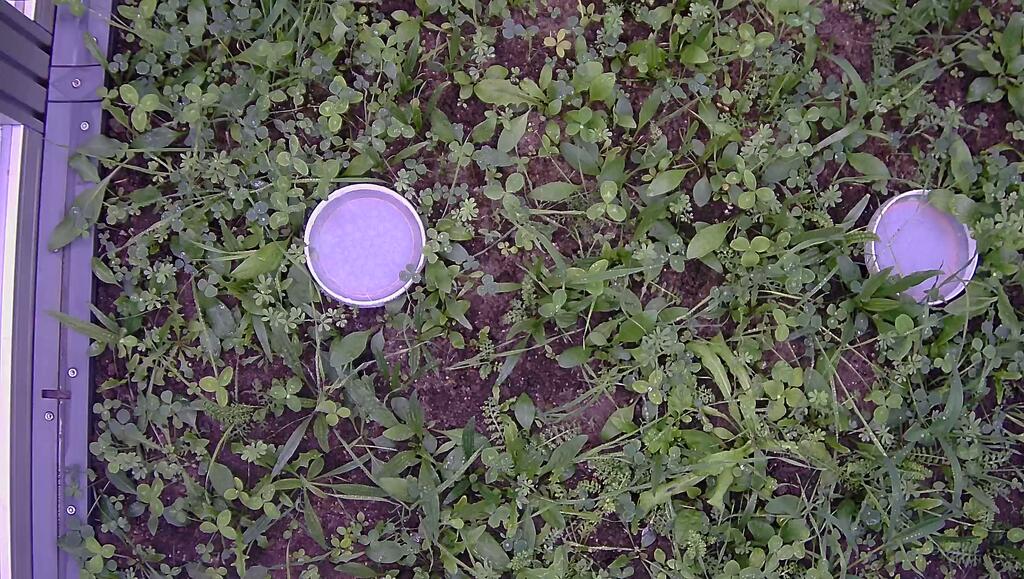}\\
		(2)\\
		Not Annotated\\[0.5em]
		\noann{}
	}
	\adjustbox{cfbox=red 1pt 1pt 0pt, minipage=0.23\textwidth}{
		\centering
		\includegraphics[width=\textwidth]{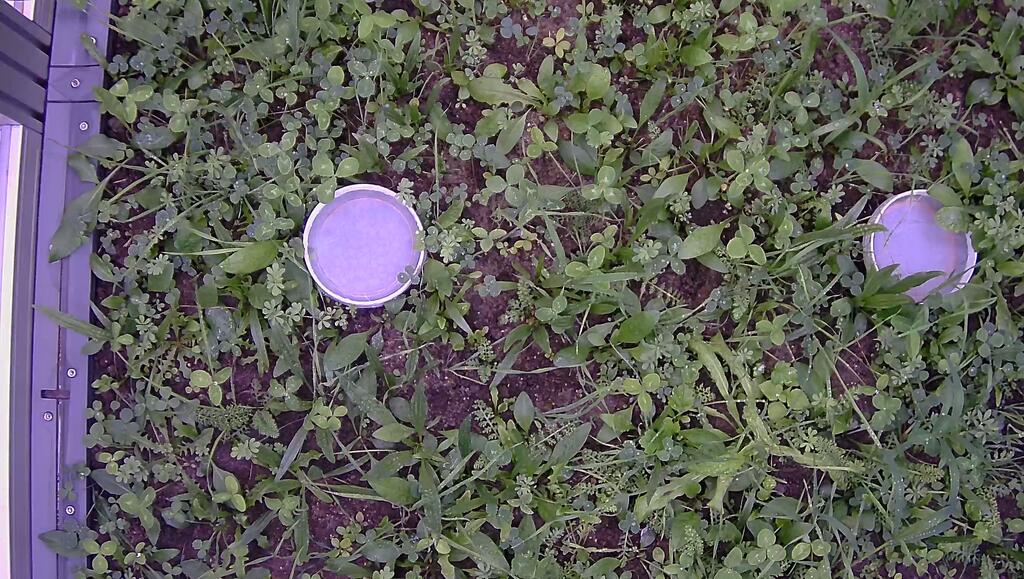}\\
		(3)\\
		Not Annotated\\[0.5em]
		\noann{}
	}\\
	\adjustbox{cfbox=red 1pt 1pt 0pt, minipage=0.23\textwidth}{
		\centering
		\includegraphics[width=\textwidth]{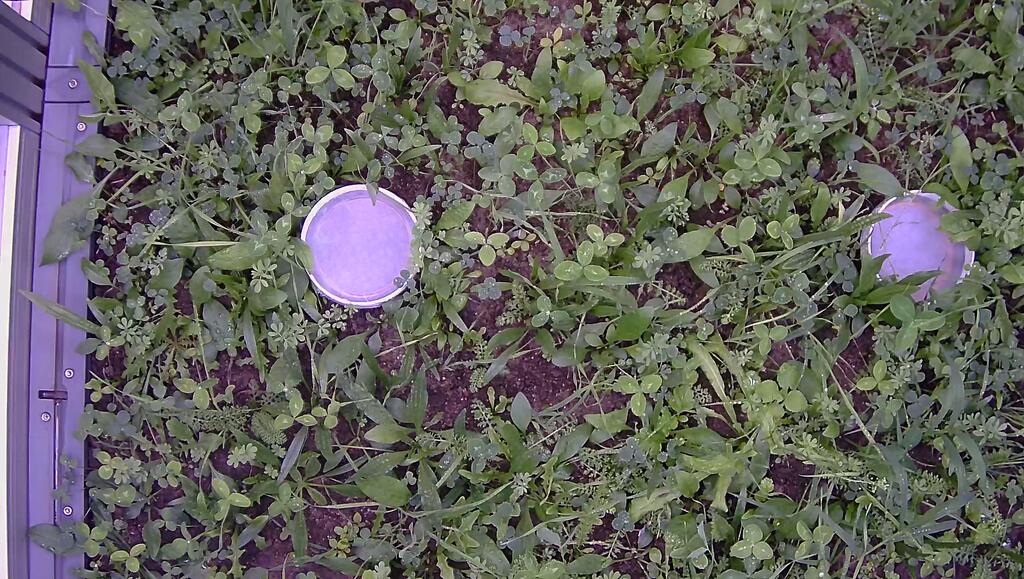}\\
		(4)\\
		Not Annotated\\[0.5em]
		\noann{}
	}
	\adjustbox{cfbox=red 1pt 1pt 0pt, minipage=0.23\textwidth}{
		\centering
		\includegraphics[width=\textwidth]{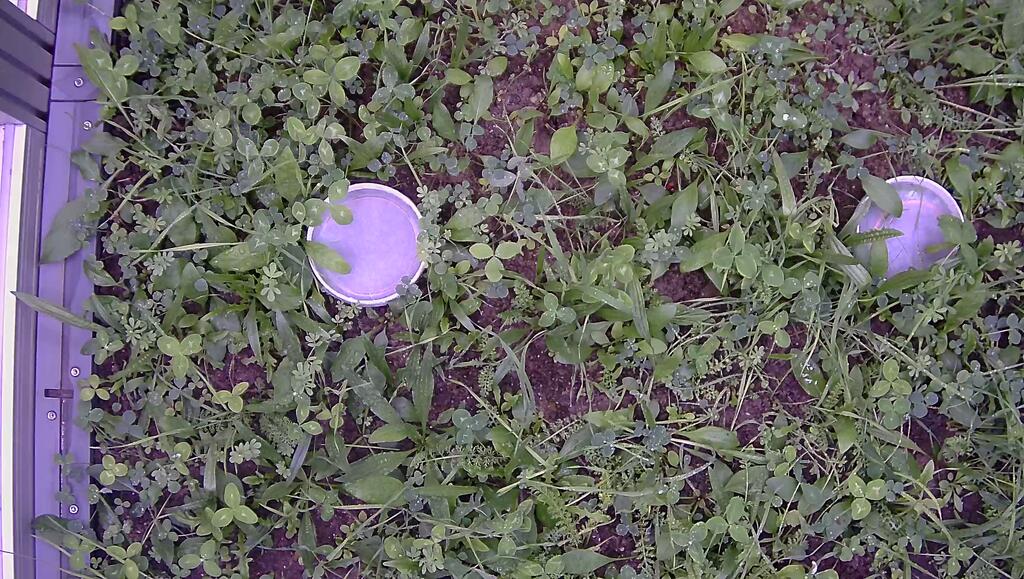}\\
		(5)\\
		Not Annotated\\[0.5em]
		\noann{}
	}
	\adjustbox{cfbox=red 1pt 1pt 0pt, minipage=0.23\textwidth}{
		\centering
		\includegraphics[width=\textwidth]{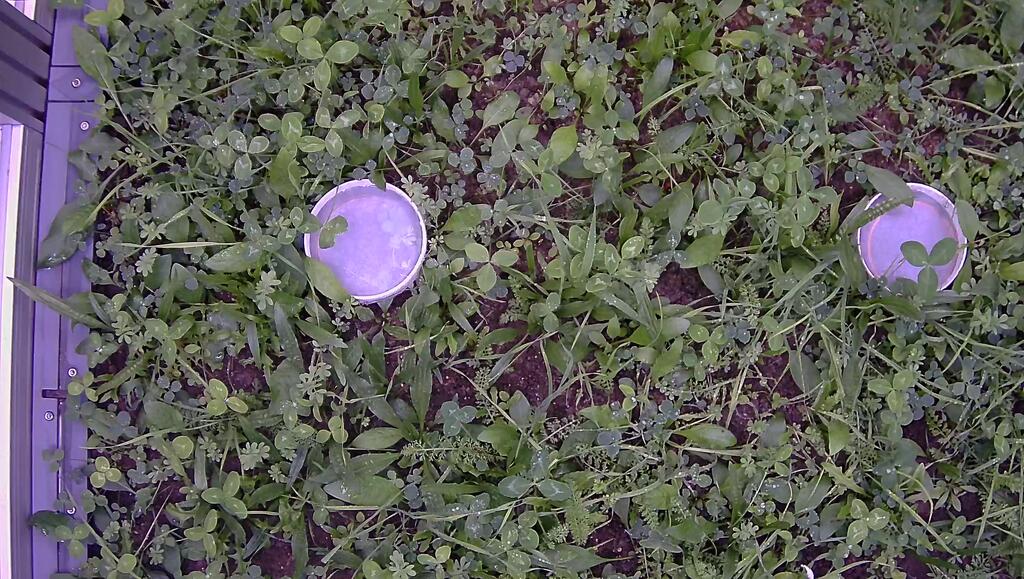}\\
		(6)\\
		Not Annotated\\[0.5em]
		\noann{}
	}
	\adjustbox{cfbox=green 1pt 1pt 0pt, minipage=0.23\textwidth}{
		\centering
		\includegraphics[width=\textwidth]{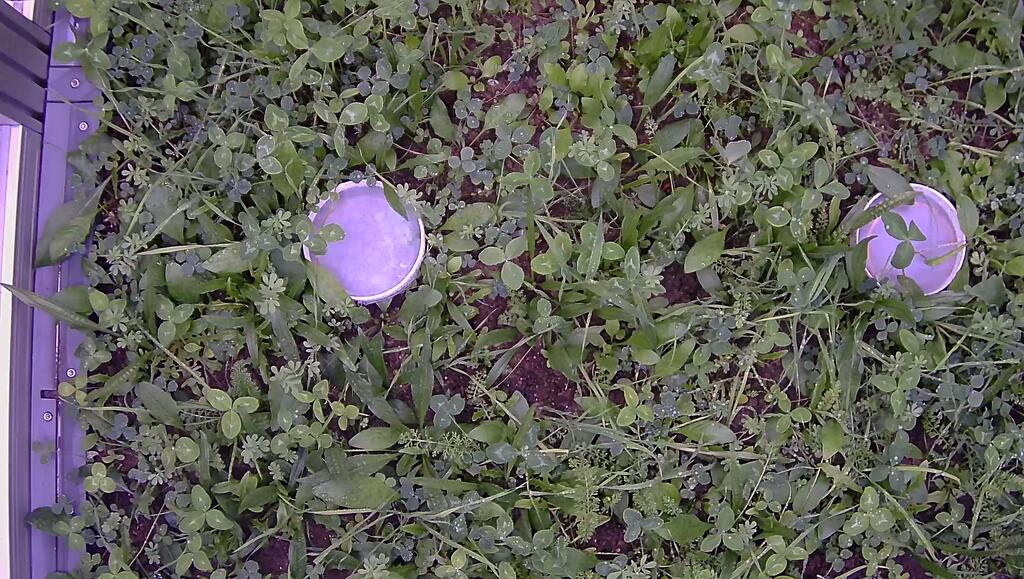}\\
		(7)\\
		Annotated\\[0.5em]
		\ann{25\%}{10\%}{40\%}
	}
	\caption{From the entire dataset, since there are weekly annotations, but daily images, only about one in seven images is annotated. While the images do not significantly differ from one day to the next, the small differences between the images can still help the model learn intermediate growth stages of the plants.}
	\label{fig:timeseries}
\end{figure}

The plant community composition is an essential indicator for environmental changes such as changes in climate change \cite{rosenzweig2007assessment,liu2018shifting,lloret2009plant}, insect abundance \cite{Souza2016BottomupAT,ulrich2020invertebrate}, and land-use \cite{gerstner2014landuse,helm2019recovery}. Hence, this kind of data is usually collected by plant ecologists \cite{liu2018shifting,gerstner2014landuse,Souza2016BottomupAT,bruelheide2018global}, for example, in the form of measuring the plant cover, in regular, but rather long, time intervals. The plant cover is defined as the percentage of area of ground covered by each plant species disregarding any occlusion. Usually, many different plant species are contained in a single plot, which often overgrow and occlude each other, making the estimation of the plant cover a very complex task.

Automated plant cover prediction can be a vital asset to plant biodiversity researchers. Abundance values, like the plant cover, are traditionally collected manually by estimating them directly in the field on vegetation plots by visual inspection (see \autoref{fig:timeseries}). However, collecting data this way is laborious, prone to human error, and subjective. Therefore, automated systems performing such estimations offer a significant advantage to these traditional methods, as they can analyze a large number of images of such vegetation plots in a short amount of time and deliver valuable research data at a high temporal resolution.
The collected plant abundance data can then be used to determine the influence of environmental changes on the plant communities. The high temporal resolution of the automatically extracted data offers the potential for very fine-grained analyses, such that a shift in the community distribution can be investigated in intervals of days or even hours instead of only weeks \cite{ulrich2020invertebrate}, months \cite{andrzejak2022effects} or years \cite{helm2019recovery}.

To establish an automated system to perform such an analysis of images, convolutional neural networks (CNNs) are a good choice, as they are powerful image processing models. However, they usually require large amounts of labeled training data to perform well. Körschens \etal \cite{koerschens2021weakly} demonstrated a way to determine the plant cover by training on the so-called InsectArmageddon dataset \cite{ulrich2020invertebrate,koerschens2021weakly}, which we will also investigate in this work and which contains merely 682 labeled images, collected and annotated with plant cover estimates in weekly intervals. As this number of training images is relatively low, especially in conjunction with such a complex task, the quality of the results is very likely limited by the amount of available training data. 

To solve the issue of little labeled data, we investigate using unlabeled intermediate images to increase the size of the training set. Plant cover estimates for vegetation plots are very laborious to create. However, additional unlabeled images are not. If an automated camera system exists to gather images for training or automatic analysis of plant cover, it can usually also collect a large number of additional unlabeled images at almost no cost. In the InsectArmageddon dataset, images are collected at a daily basis but only annotated at the weekly one. We leverage this experimental setup to automatically generate weak labels for the intermediate days between two days for which human annotations are given. The key idea is to handle uncertainty in the weak labels by weighting them according to their temporal distance to the next reference estimate. We will refer to this approach as label interpolation.



In addition to this, to enable the network to train on images at their full resolution, we propose a Monte-Carlo sampling approach for training the network, which we will refer to as Monte-Carlo Cropping (MCC). The original image is sampled in equally-sized patches, for each of which the target output is estimated individually. Afterwards, the network output for all patches sampled from a single image is averaged. This kind of sampling empirically seems to have a regularizing effect on the network training, leading to better results on high-resolution images and drastically reducing the training time of such images.

In the following, we will elaborate on related work to our approaches, followed by a detailed explanation of our methods, experimental results, and finally, a conclusion.

\section{Related Work}


\paragraph{Label Interpolation.}

In this work, we investigate the problem of utilizing unlabeled images for training in addition to a small number of images with labels. This problem is usually tackled by semi-supervised learning approaches, especially self-training \cite{scudder1965probability,kahn2020self}. In self-training methods, a model is trained with few annotated images and then used to label available but unannotated images to increase the size of the training set iteratively. In contrast to these approaches, label interpolation heavily utilizes the strong correlation of plant cover values in the time series to generate labels and does not rely on trained models at all.
Similarly, the data augmentation method mixup \cite{zhang2017mixupBE} also interpolates labels to generate novel annotations. However, in contrast to our method, the authors do not apply the new labels to unlabeled images but fuse two existing images and their class labels.

\paragraph{Monte-Carlo Cropping (MCC).}

Another problem we tackle in this work is the utilization of high-resolution images in CNN training. Cropping the original training images into much smaller images is a simple approach to this problem, and is usually applied in tasks like image segmentation and object detection, that also often deal with high-resolution images \cite{wang2022internimage,yuan2019segmentation,cheng2020panoptic,li2022efficient,li2023yolov6}. For these tasks, cropping is usually done a single time per epoch per image, and the ground-truth data is also adapted in the same way. For image segmentation, the ground-truth data are usually segmentation maps, which have the same dimensions as the original image, and can also be cropped in the same way. The ground-truth for object detection are usually bounding box coordinates in the original images, which can also be easily be adapted to the cropped input image by systematically modifying the coordinates. For plant cover estimation, however, the target data are merely numerical vectors representing the plant cover distribution in the image, and can therefore neither be cropped or simply adapted. To solve this problem, our Monte-Carlo Cropping introduces a stochastical component in order to be able to approximate the underlying plant cover distribution, which does not need to be done for image segmentation and object detection.


\section{Methods}


\subsection{Label Interpolation}

\newcommand{\cov}{\ensuremath{\text{cover}}}

The first method we introduce is label interpolation. As shown in \autoref{fig:timeseries}, from seven existing images in a single week, only a single one is labeled, leaving the other images unused. Moreover, we can see that the differences between the daily pictures are only minor compared to the weekly differences; however not insignificant. Images of this kind have two advantages. Firstly, the network can learn the growth process of plants in much more fine-grained steps, especially since this kind of data contains more and new information compared to simple augmented images. And secondly, since the differences between the pictures are relatively small, we can infer certain properties of the supposed labels for these images from their neighboring annotations.
More formally, for our label interpolation method to work, we take advantage of the fact that plant cover estimates are continuous values. Moreover, we assume that the intermediate value theorem \cite{bolzano1810beytrage} holds for these estimates collected in a time series. That is, if a plant's measured cover value in a certain week was $\cov(t_0)$ and $\cov(t_1)$ in the following week,
\begin{gather}
	\forall v \text{ with }\min(\cov(t_0), \cov(t_1)) < v < \max(\cov(t_0), \cov(t_1)), \\
	\exists t \in (t_0, t_1): \cov(t) = v\:.
\end{gather}
Under the assumption that plants grow in a continuous fashion without external interference, this theorem holds.

Here, we will utilize linear interpolation, specifically of the data of two subsequent weeks, which implicitly weights the values respective to their temporal distance to the next annotated data point:
\begin{equation}
	\cov(t) = \frac{\cov(t_0)(t_1 - t) + \cov(t_1)(t - t0)}{t_1 - t_0}\:.
\end{equation}

A linear interpolation might not precisely represent the growth process of the plants and discontinuities in the images (like occlusion). However, with such small time steps, the growth process of the plants can be assumed to be approximately linear between two weeks. We are aware of violations of our assumptions in practice. However, empirically such issues play only a minor role when it comes to the overall quality of our suggested approach. 



\subsection{Monte-Carlo Cropping}

The second method we introduce here is Monte-Carlo Cropping (MCC). A significant problem in plant cover prediction is that the images provided by the camera systems usually have a relatively high resolution (e.g., $2688\times 1520$ pixels for the InsectArmageddon dataset). However, networks are usually only applied on rather small, often downscaled images ($224\times 224$ for typical ImageNet \cite{russakovsky2015imagenet} tasks, and $448\times 448$ or similar for fine-grained ones \cite{cui2018large}).

Training on large images is computationally expensive, consumes large amounts of memory, and can take a long time. For the original image $I \in \mathbb{R}^{H\times W \times 3}$ we sample patches $P \in \mathbb{R}^{h\times w \times 3}$ with $h \ll H$ and $w \ll W$. For each patch $P$, we let the network predict the plant cover separately and then average these values over the number of patches sampled from each image.

Since the patches are sampled from an image with the plant cover values $\cov_p$, the expected value is equal to $\cov_p$ for a large number of patches sampled. Therefore, due to the law of large numbers \cite{evans2004probability}, the following holds:

\begin{equation}
	\lim_{n \rightarrow \infty} \frac{1}{n}\sum_{i=1}^n \cov_{i,p} = \cov_p\:,
\end{equation}%
with $n$ being the number of patches sampled, $i$ being the index of the randomly sampled patch, and $p$ denoting plant species. I.e., while the plant cover values observed in the smaller patches do not necessarily reflect the values of the total images when selecting a sufficiently large number of patches, they do so on average.




\begin{figure}[t]
	\centering
	\includegraphics[width=0.49\textwidth]{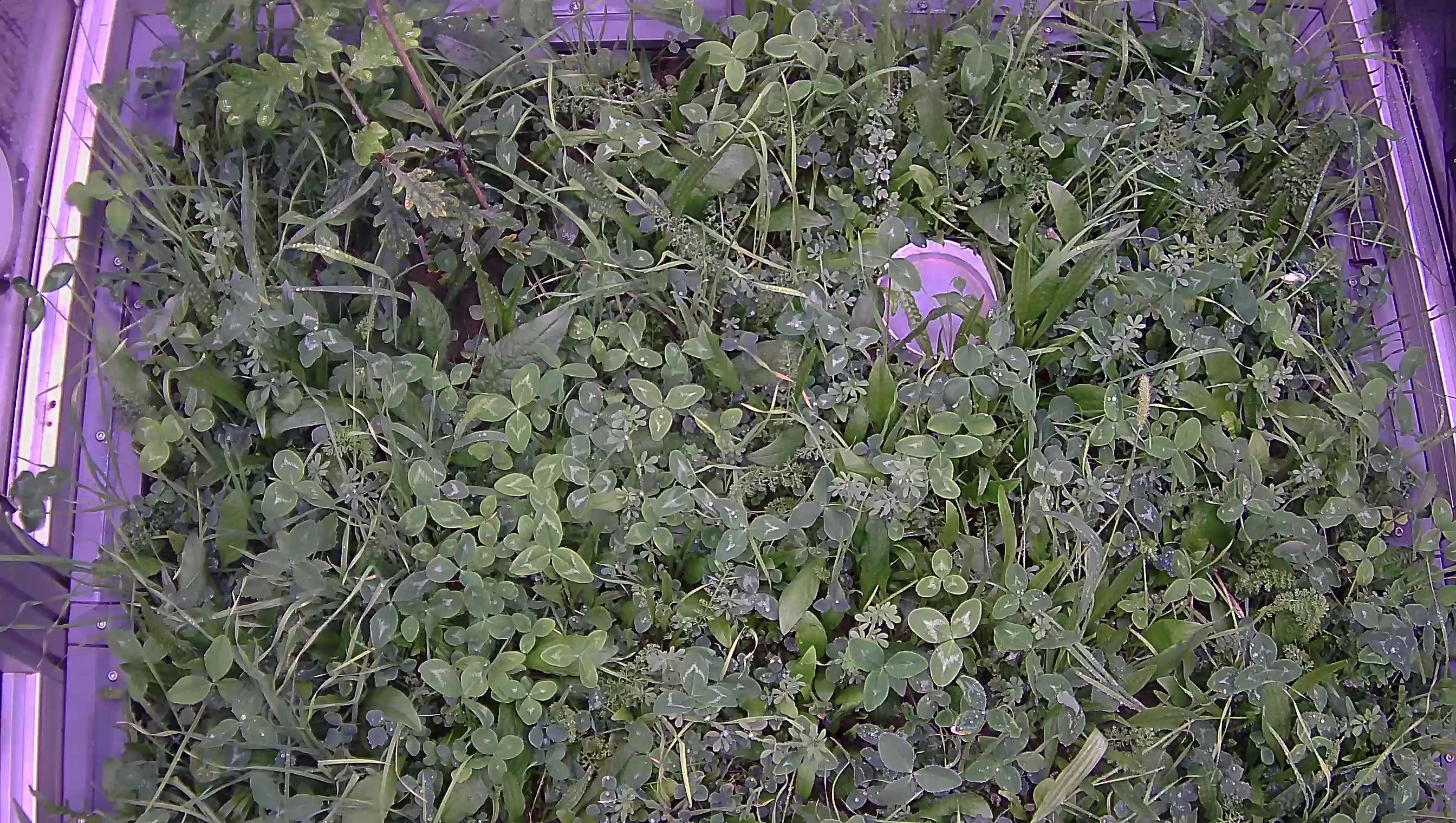}
	\hfill
	\includegraphics[width=0.49\textwidth]{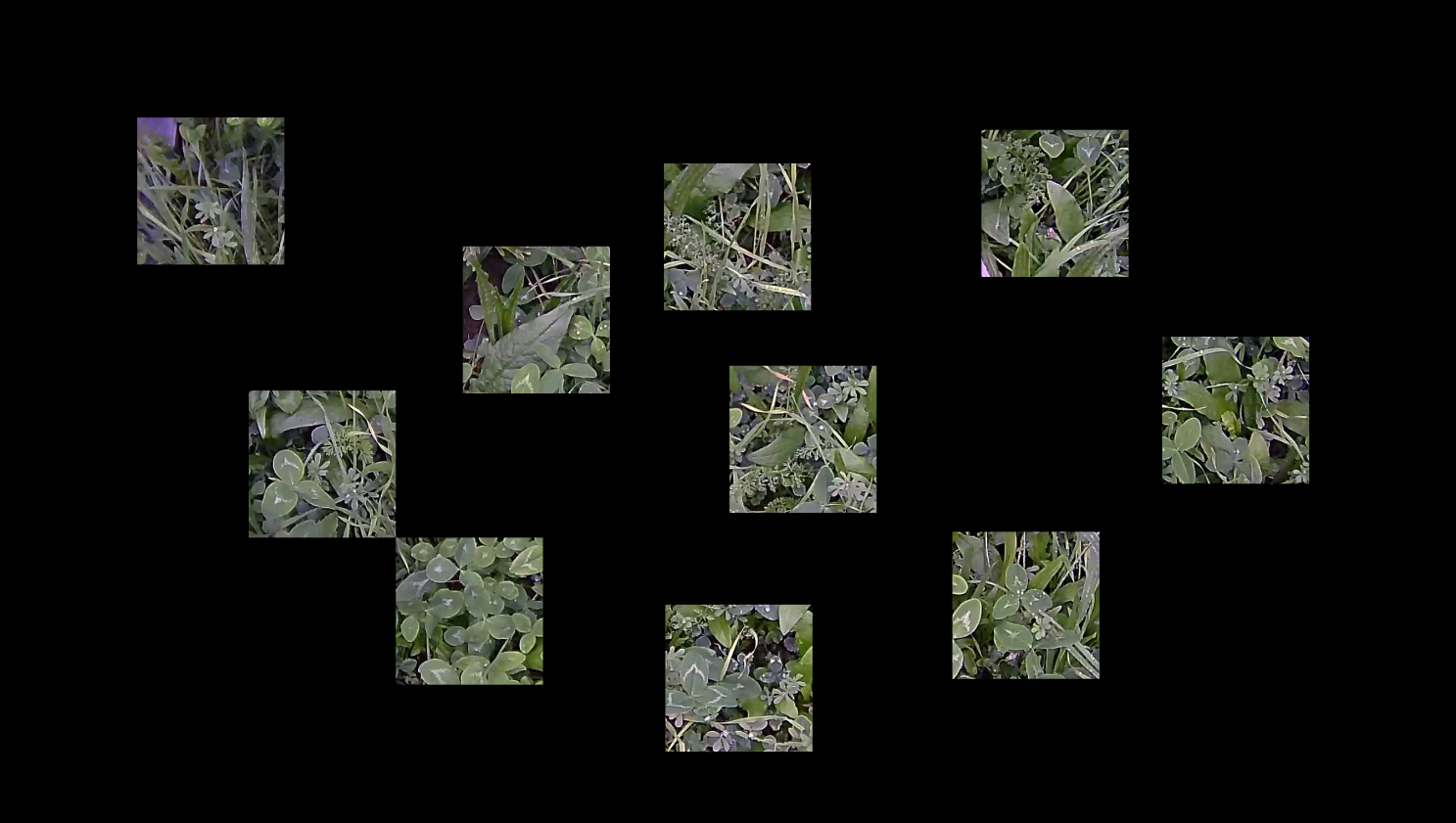}
	\caption{An example of a random patch selection with Monte-Carlo Cropping.}
	\label{fig:mc_orig_comparison}
\end{figure}

During training, we sample equally sized square patches from the original image, an example of which can be seen in \autoref{fig:mc_orig_comparison}. It is visible that the number of pixels shown to the network is significantly reduced, depending on the size of the patches and number of patches sampled. Hence, computational complexity can also be reduced with MCC.

\section{Experimental Results}

\subsection{Dataset}

In our experiments, we utilize the InsectArmageddon dataset \cite{ulrich2020invertebrate,koerschens2020towards} from the eponymous iDiv project that took place in 2018 over multiple months. The images were collected in 24 so-called EcoUnits, which are boxes containing small enclosed ecosystems. Each of the EcoUnits was equipped with two cameras that collected daily pictures of these ecosystems. The dataset from the InsectArmageddon experiment comprises estimated plant cover data (``reference estimates'') for eight herbaceous plant species in 682 images collected weekly by a single ecologist. The image set with original and interpolated annotations contains about 4900 images, i.e., about seven times the number of images due to one originally labeled image per week, and six with interpolated labels. On this dataset, we perform 12-fold cross-validation by selecting the images of two EcoUnits for testing, and the ones of the remaining 22 EcoUnits for training. For more details on the InsectArmageddon image dataset, we would like to refer to \cite{ulrich2020invertebrate} and \cite{koerschens2020towards}.

\subsection{Setup}

\newcommand{\msae}{MSAE$_\sigma$}

We use the approach introduced in \cite{koerschens2021weakly}. i.e., we utilize a ResNet50 \cite{he2016deep}, architecture with Feature Pyramid Network \cite{lin2017feature}, the 3-phase pre-training pipeline based on freely available images from GBIF\footnote{\url{http://gbif.org}} to train a classification network (phase 1), which generates simple segmentations with class activation mapping (CAM) \cite{zhou2016cam}, on which we then pre-train a segmentation network (phase 2). The weights of this network are then used as initialization for our plant cover prediction network, which we train on the plant cover annotations (phase 3).

During the first phase, we utilize global log-sum-exp-pooling \cite{koerschens2022beyond,koerschens2021weakly} with a learning rate of $10^{-4}$ and a weight decay of $10^{-4}$. We use this pooling method, as in \cite{koerschens2022beyond} it generated better segmentations when used in conjunction with CAM. Moreover, we use a categorical cross-entropy loss during optimization and train with early stopping and dynamic learning rate reduction. The learning rate is reduced by a factor of 10 when there is no improvement in the validation accuracy over four epochs, and the training is stopped, if there is no improvement over six epochs.
In the second phase, we use a learning rate of $10^{-5}$, a weight decay of $10^{-4}$, and a combination of binary-cross-entropy and dice loss, which are summed up and weighted equally as loss. During this training, we also used a dynamic learning rate adaptation and early stopping; however, we monitored the mean Intersection over Union (mIoU) instead of the accuracy.
In the third phase, we train with a batch size of 1 and a learning rate of $10^{-5}$, which is reduced by a factor of 10 after 50\% and 75\% of the total epochs, respectively. We are using the mean scaled absolute error (\msae{}) as loss. This error is defined as

\begin{equation}
    MSAE_{\bm{\sigma}}(\bm{t}, \bm{p}) = \frac{1}{n} \sum_{i=1}^n \left|\frac{t_i}{\sigma_i} - \frac{p_i}{\sigma_i}\right|\:,
\end{equation}

where $\sigma$, in our case, is the standard deviation of the species-wise plant cover values calculated over the training dataset. This loss aims to reweight the species to account for the substantial imbalance in the dataset.

In our experiments, we compare the image resolution used in previous works \cite{koerschens2021weakly} ($1536\times 768$ pixels) and the full image resolution ($2688\times 1536$ pixels). We investigate several training durations and their effect on the training with weekly and interpolated daily images. For the daily labels we investigate fewer epochs, since the iteration count per epoch is much higher in comparison to the weekly label dataset.

For our experiments on MCC, we investigate different patch sizes as well as several patch counts. We chose the patch sizes of $128^2$, $256^2$, $512^2$, and $1024^2$ pixels, and for each patch size, a respective sample count so that the number of pixels sampled is around 50\% of the pixel count of the original image. For each sample count selected, we also investigate a pixel count of half or double the number of pixels sampled. For a patch of $512^2$ pixels, we investigate sample counts of 8, 4, and 16; for a patch of $256^2$ pixels, sample counts of 32, 16, and 64, etc. It should be noted that the cropped image patches are input into the network as-is without any additional resizing.

\subsection{Metrics}

We investigate three different metrics to analyze our results. The first metric is the aforementioned mean scaled absolute error MSAE$_\sigma$. The second metric is the mean Intersection over Union (IoU) metric calculated over the segmentation image subset introduced in \cite{koerschens2021weakly} containing 14 pixel-wise annotated images from the InsectArmageddon dataset. The last metric we will refer to as the DCA-Procrustes-Correlation (DPC). It is calculated by performing a Detrended Correspondence Analysis (DCA) \cite{hill1980detrended} on the target and predicted outputs, which are then compared with a Procrustes analysis. This returns a correlation value, where higher values show a higher similarity of the distributions to each other, which is significant for ecological applications.

With the \msae{}, we can evaluate the performance of our models in absolute terms, i.e., how accurate the species-wise predictions are based on the reference estimates. The IoU determines how well the top layer of plants is predicted, disregarding any occluded plants, and the DPC explains how well the predicted species distribution matches with the one estimated by the expert. All experiments are performed in a 12-fold cross-validation over three repetitions.

\subsection{Label Interpolation}
\label{sec:exp_label_interpolation}

\begin{table}[t]
	\centering
	\caption{Comparison of training with weekly images with only the original labels and daily images with original and interpolated labels. Abbreviations used: \msae{} - Mean Scaled Absolute Error, IoU - Intersection over Union, DPC - DCA-Procrustes-Correlation. Top results are marked in \textbf{bold font}.}
	\begin{tabular}{ll|ccc|ccc}
\toprule
          &    & \multicolumn{3}{|c}{Weekly Images} & \multicolumn{3}{|c}{Daily Images} \\
          &    &          \msae{} &   IoU &   DPC &         \msae{} &   IoU &   DPC \\
\textbf{Resolution} & \textbf{Epochs} &               &       &       &              &       &       \\
\midrule
\multirow{6}{*}{\textbf{1536x768}} & \textbf{3 } &         0.527 & 0.158 & 0.724 &        0.499 & 0.198 & 0.780 \\
          & \textbf{6 } &         0.505 & 0.192 & 0.766 &        0.502 & 0.204 & 0.766 \\
          & \textbf{10} &         0.501 & 0.196 & \textbf{0.770} &        0.503 & 0.199 & 0.772 \\
          & \textbf{15} &         0.500 & 0.201 & 0.768 &        0.498 & 0.194 & 0.773 \\
          & \textbf{25} &         0.501 & 0.203 & 0.760 &            - &     - &     - \\
          & \textbf{40} &         0.502 & 0.187 & 0.765 &            - &     - &     - \\
\cline{1-8}
\multirow{6}{*}{\textbf{2688x1536}} & \textbf{3 } &         0.545 & 0.156 & 0.656 &        0.494 & 0.205 & 0.780 \\
          & \textbf{6 } &         0.510 & 0.188 & 0.757 &        0.493 & \textbf{0.223} & 0.778 \\
          & \textbf{10} &         0.502 & 0.204 & 0.766 &        0.491 & 0.208 & 0.777 \\
          & \textbf{15} &         0.497 & \textbf{0.208} & 0.757 &        \textbf{0.489} & 0.181 & \textbf{0.781} \\
          & \textbf{25} &         \textbf{0.493} & 0.205 & 0.763 &            - &     - &     - \\
          & \textbf{40} &         0.495 & 0.192 & 0.761 &            - &     - &     - \\
\bottomrule
\end{tabular}

	\label{tbl:label_interpolation}
\end{table}

The results of our experiments with label interpolation are shown in \autoref{tbl:label_interpolation}. Regarding the weekly annotated images, it is visible that the \msae{} and IoU are increasing with higher epoch counts, and the higher-resolution images also return slightly better results than the low-resolution images. However, after ten epochs, low-resolution images achieve the best DPC value (0.77). This value is not outperformed when using high-resolution images, leading to the conclusion that the network can learn to reduce the prediction error for some more dominant species from the high-resolution images but cannot accurately learn and reflect the actual distribution due to the neglect of less abundant species.

When looking at the results of the experiments using the interpolated daily images in conjunction with the weekly annotations, we notice improvements for both image resolutions, showing that our interpolation method is effective and leads to better results than just using the annotated images. The low-resolution and high-resolution images with daily images outperform their counterparts in all metrics. It should also be noted that the top performance is achieved after a smaller number of epochs for the daily images, likely because the number of iterations per epoch is about seven times the one with weekly images. This way, the top DPC value is achieved after three epochs, while the top IoU is achieved after six epochs. The \msae{} appears to be still improving for a larger number of epochs.

\subsection{Monte-Carlo Cropping}



\newcommand{\sub}[2]{
	\begin{subfigure}{0.48\textwidth}
		\centering
		\includegraphics[width=\textwidth]{#1}
		\caption{#2}
	\end{subfigure}
}

\begin{figure}
	\centering
	\begin{tabular}{cc}
		Weekly Images & Daily Images \\
		\sub{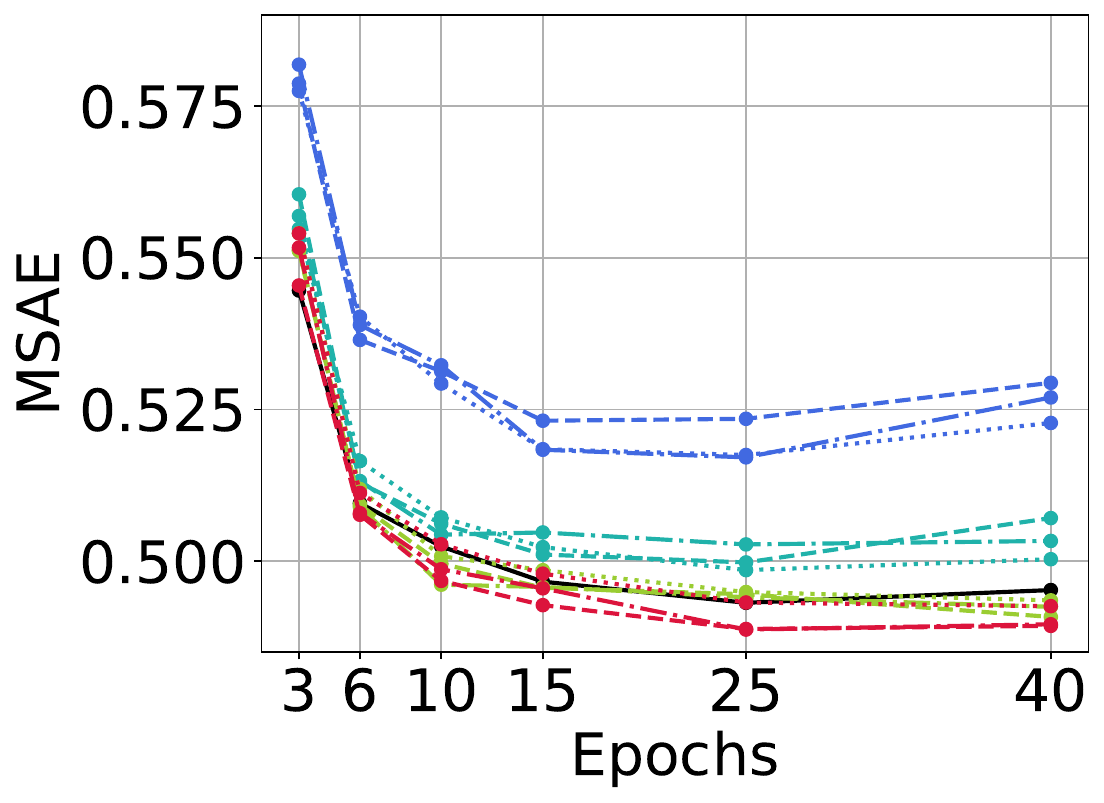}{\msae{}} &
		\sub{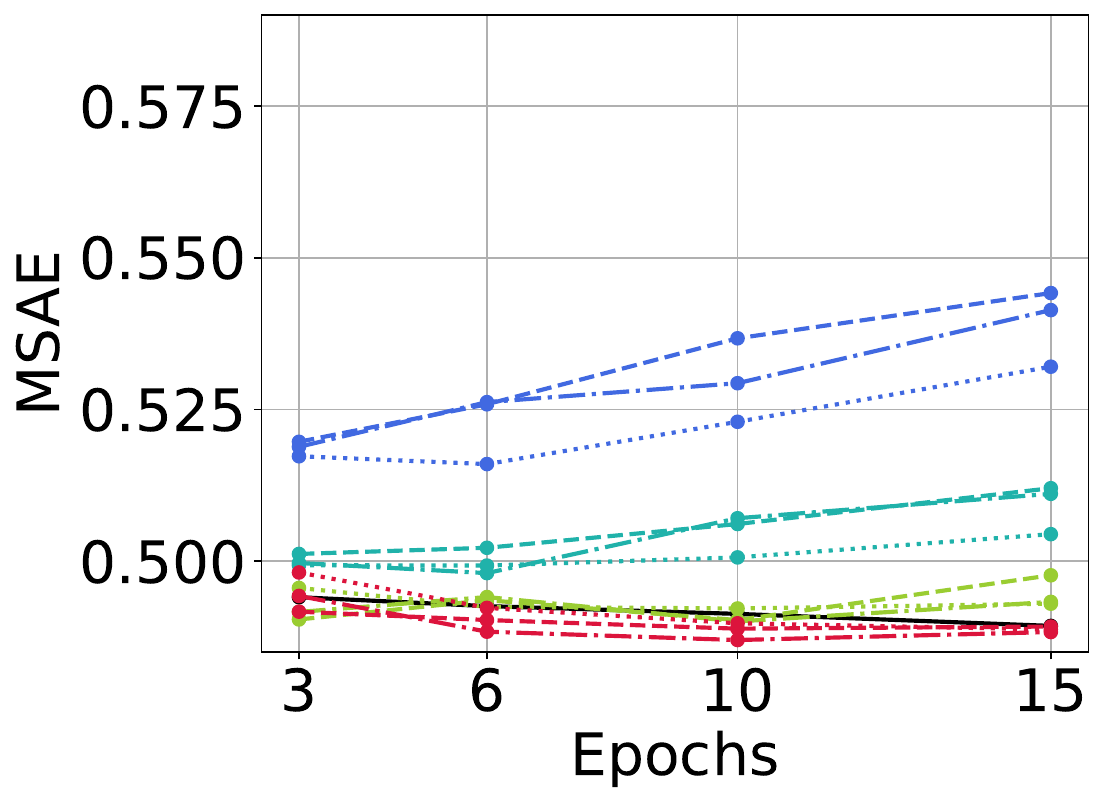}{\msae{}} \\
		\sub{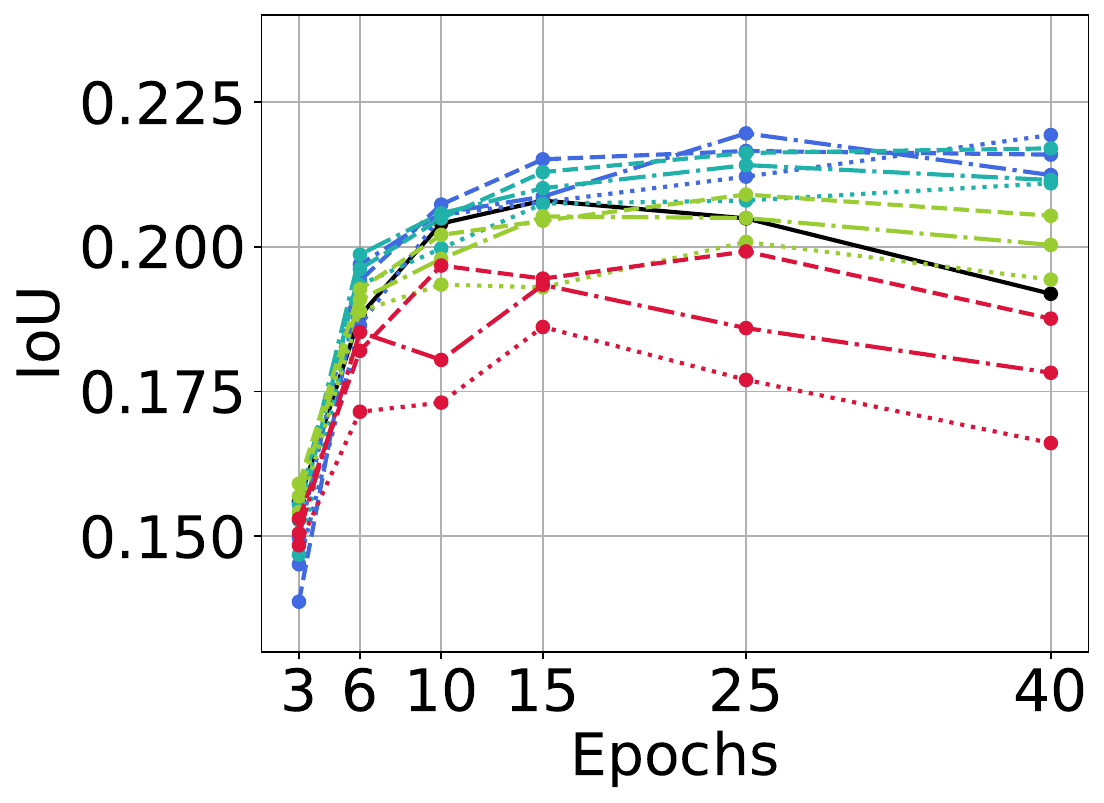}{IoU} &
		\sub{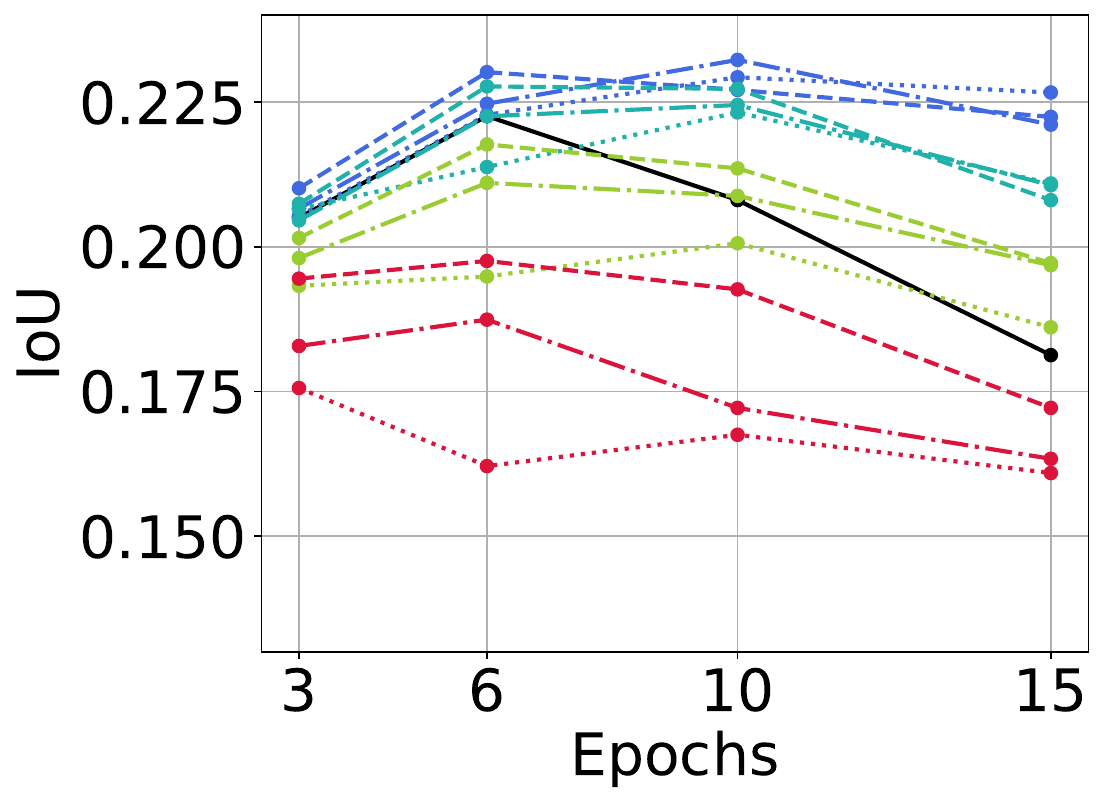}{IoU} \\
		\sub{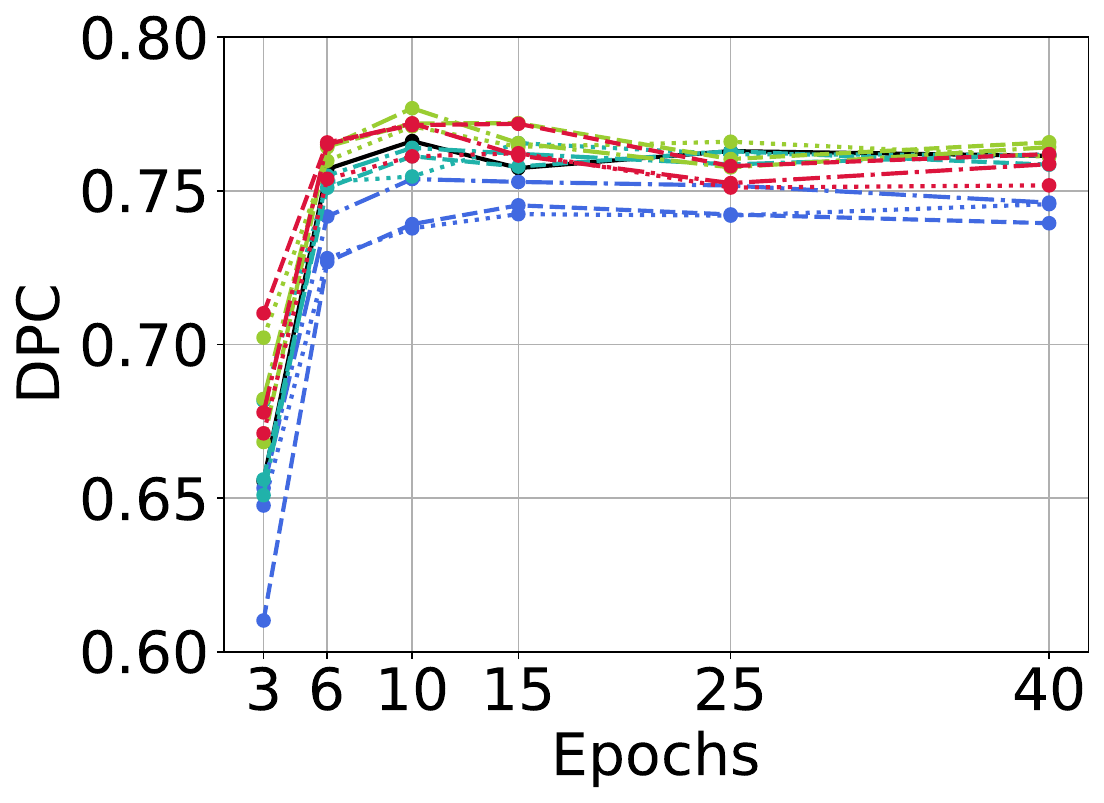}{DPC} &
		\sub{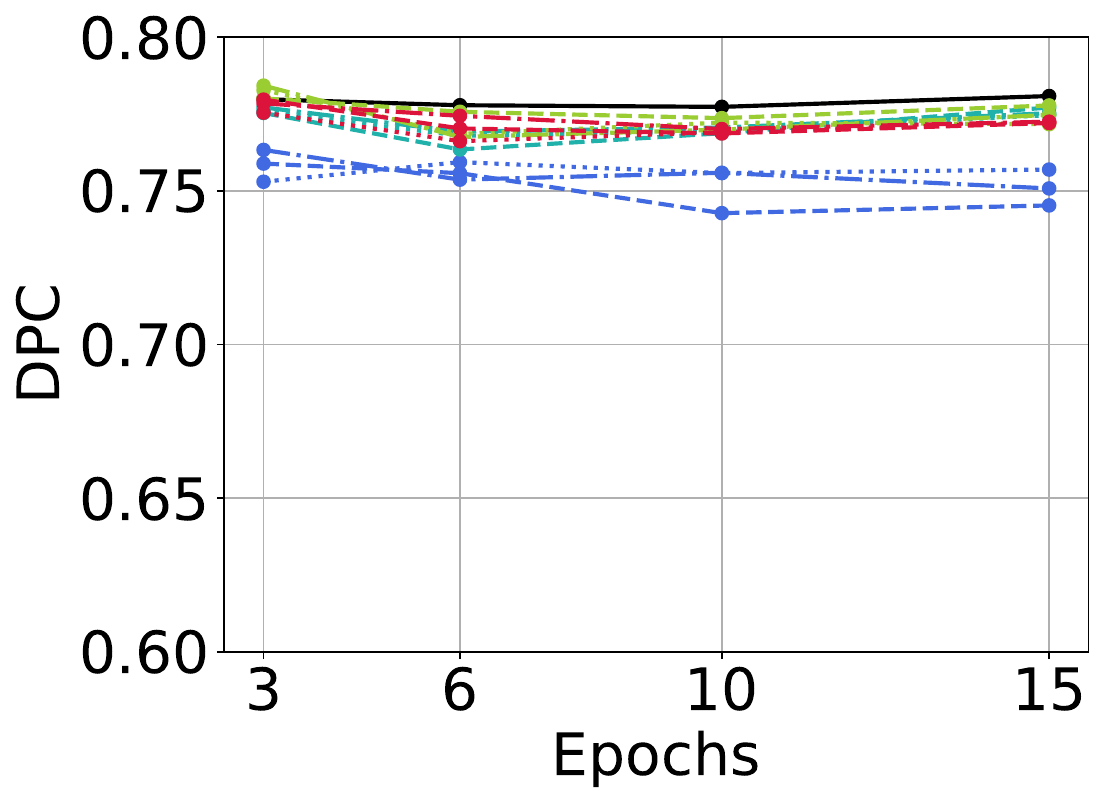}{DPC} \\
		\multicolumn{2}{c}{
			\centering
			\includegraphics[width=0.9\textwidth]{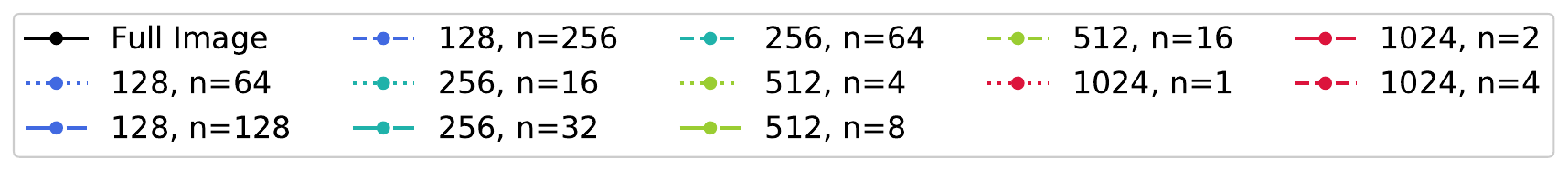}
		} \\
	\end{tabular}
	\caption{The development of the different metrics over several training durations for images with weekly labels (left) and images with interpolated (daily) labels (right). Abbreviations used: \msae{} - Mean Scaled Absolute Error, IoU - Intersection over Union, DPC - DCA-Procrustes-Correlation}
	\label{fig:plot_weekly_daily}
\end{figure}

The results of our experiments with MCC on full-resolution images with weekly labels and interpolated daily labels are shown in \autoref{fig:plot_weekly_daily}. The detailed numerical results can be found in the Supplementary Material. Generally, we can see that larger patch sizes lead to better results in terms of \msae{} and DPC, with the patch sizes 512 and 1024 yielding the top results for the experiments with daily and weekly labels. The patch size of 512 yields the best results for DPC, while the size of 1024 performs best in terms of \msae{}. For the weekly labels, the top DPC value and \msae{} value are 0.777 and 0.489, which outperform the top results on the full-resolution weekly images with a DPC of 0.766 and \msae{} of 0.493, respectively. For achieving this performance, for the patch size 512 the optimal sample size is 8, and for 1024 it is 2, representing about 50\% of the number of pixels of the original image. The same configurations generate the best \msae{} and DPC results for the daily images, with 0.487 and 0.784, respectively. The MC training outperforms the full image training also here in terms of \msae{} (0.487 vs. 0.489) and DPC (0.784 vs. 0.781).

Interestingly, the best top layer prediction results, i.e., segmentation results, were achieved with a much smaller patch size, of $128^2$ pixels. For the weekly images and a sample size of 128, the top IoU was 0.220, again outperforming the naive full-resolution approach with an IoU of 0.208. Similarly, with a patch size of 128 and a sample size of 128, the training on the daily images yields an IoU outperforming the full-resolution training (0.232 vs. 0.223).

The discrepancies between the configurations for optimal IoU and optimal \msae{} and DPC can be explained by what kind of features the network learns for the different patch sizes. With smaller patches, the network is forced to focus on the single plants shown in the top layer and learns little about the relationships of plants between each other, like occlusion. These relationships, however, play a significant role in the accurate prediction of the species-wise values and the entire composition, which are evaluated by \msae{} and DPC, respectively. Larger patch sizes capture the relationships and therefore perform better in these aspects.

In summary, our MCC approach can outperform the training with full-resolution images. The results significantly depend on the patch size, with higher patch sizes resulting in better community-based predictions and smaller ones in better individual-based predictions. As training on smaller patches instead of a large image has computational implications, we will compare computation times in the following.

\paragraph{Computation Time Comparison}

\begin{table}[t]
	\centering
	\caption{Comparison of training speed per epoch on the full resolution images using different patches and sample sizes. Times shown are in minutes:seconds.}
	\begin{tabular}{cc|cc}
\toprule
     &   &  Weekly Images &   Daily Images \\
     &   & Time per Epoch & Time per Epoch \\
\textbf{Patch Size} & \textbf{\#Patches} &                &                \\
\midrule
\textbf{-} & \textbf{-} &          01:51 &          16:17 \\
\cline{1-4}
\multirow{3}{*}{\textbf{128}} & \textbf{64} &          00:55 &          09:06 \\
     & \textbf{128} &          01:34 &          14:01 \\
     & \textbf{256} &          03:01 &          24:25 \\
\cline{1-4}
\multirow{3}{*}{\textbf{256}} & \textbf{16} &          00:46 &          08:22 \\
     & \textbf{32} &          01:08 &          10:37 \\
     & \textbf{64} &          02:10 &          18:22 \\
\cline{1-4}
\multirow{3}{*}{\textbf{512}} & \textbf{4} &          00:46 &          08:07 \\
     & \textbf{8} &          01:03 &          10:16 \\
     & \textbf{16} &          01:55 &          16:41 \\
\cline{1-4}
\multirow{3}{*}{\textbf{1024}} & \textbf{1} &          00:45 &          08:15 \\
     & \textbf{2} &          01:02 &          10:23 \\
     & \textbf{4} &          01:52 &          16:17 \\
\bottomrule
\end{tabular}

	\label{tbl:mc_speedcomp}
\end{table}

A comparison of the times per epoch for each setup using the full-resolution images is shown in \autoref{tbl:mc_speedcomp}. These measurements were taken when training on about 75\% of the images of the weekly and daily image sets on an RTX 3090.
The training on the original full-resolution images took 1 minute and 51 seconds per epoch for only the weekly images and 16 minutes and 17 seconds for daily images. The top results for \msae{} and DPC were generated by patch size 512 and 8 patches sampled. This setup takes 1 minute and 3 seconds on the weekly images and 10 minutes and 16 seconds on daily ones, resulting in a time reduction of about a third. As the setup with patch size 512 and 4 sampled patches performs comparably well, one could even reduce the training time further by about 50\%, at little cost in performance.
The setup generating the top segmentation results, i.e., patch size 128 and sample sizes 256 and 128, differ in the training durations. Considering the larger numbers of pixels used for sample size 256, the duration is longer for this setup, requiring about 63\% more time for an epoch. In contrast, the sample size of 128 reduces the training time again by about 14\%. It should be noted that, with MCC, the number of epochs required for the optimal results on the weekly images are similar to the number of epochs for full-resolution training, and on the daily images the training times are usually even shorter for MCC training. For example, with MCC training on daily images the best DPC value is achieved after 3 epochs as opposed to 15 for full-resolution training. Hence, the training times are not only reduced regarding the per-epoch duration, but also the number of epochs in total.

\section{Conclusion \& Future Work}

We introduced two approaches for improving the data efficiency for plant cover estimation training. One method utilizes the unannotated images in the dataset, which can be collected at almost no cost; the other one enables efficient training at high resolution, gathering more information from the high resolution of the images.

Both approaches have proven effective: the label interpolation led to sufficient training data to receive improved results for all investigated metrics when using full-resolution images compared to their lower-resolution counterparts. Therefore, it is advantageous to collect more images than can be labeled, if the images are similar enough to existing images, as we can artificially increase the size of the dataset by interpolating.
Furthermore, the Monte-Carlo Cropping improved these results even further, producing better results for different aspects of the plant cover prediction task while decreasing the training time and computation required during training. While, of course, a higher image resolution contains more information, with our MCC approach, especially in combination with label interpolation, such a high resolution can be utilized much more effectively. By increasing the image resolution in future experiments even further than in the InsectArmageddon experiment, with our methods such a high resolution can actually be utilized to improve the cover estimates without additional human effort.

Our approaches will be evaluated further on new plant cover datasets with different image resolutions, more frequent images and more plant species in future work. Moreover, the label interpolation might be improved with a more sophisticated interpolation model instead of linear interpolation, for example, a model that considers the information in the image for interpolation. Similarly, the sample selection of the Monte-Carlo Cropping could be improved by intelligently selecting the patches that contain the most information in the future.

\ifreview
\else
\section*{Acknowledgements}

Matthias Körschens thanks the Carl Zeiss Foundation for the financial support. We thank Alban Gebler for enabling the image collection process in the iDiv EcoTron. 
We acknowledge funding from the German Research Foundation (DFG) via the German Centre for Integrative Biodiversity research (iDiv) Halle-Jena-Leipzig (FZT 118) for the support of the FlexPool project PhenEye (09159751).
\fi

%
%
%
\bibliographystyle{splncs04}
\bibliography{references}

\end{document}